\documentclass[10pt,twocolumn,letterpaper]{article}

\usepackage{3dv}
\usepackage{times}
\usepackage{epsfig}
\usepackage{graphicx}
\usepackage{amsmath}
\usepackage{amssymb}

\usepackage{tabulary}
\usepackage{mathtools}
\usepackage{multirow}
\usepackage{multicol}
\usepackage{bm}
\usepackage{booktabs}
\usepackage{url}
\usepackage[ruled,linesnumbered]{algorithm2e}
\DeclarePairedDelimiter{\ceil}{\lceil}{\rceil}

\usepackage[pagebackref=true,breaklinks=true,letterpaper=true,colorlinks,bookmarks=false]{hyperref}
\usepackage{appendix}
\usepackage[nocomma]{optidef}

\threedvfinalcopy 

\def\threedvPaperID{210} 

\ifthreedvfinal\fi
\begin{document}

\title{Efficient Scene Compression for Visual-based Localization}


\author{Marcela Mera-Trujillo\\
West Virginia University\\
{\tt\small mameratrujillo@mix.wvu.edu}
\and
Benjamin Smith\\
West Virginia University\\
{\tt\small bbsmith1@mix.wvu.edu}
\and
Victor Fragoso\\
Microsoft\\
{\tt\small victor.fragoso@microsoft.com}
}

\maketitle

\begin{abstract}
Estimating the pose of a camera with respect to a 3D reconstruction or scene representation is a crucial step for many mixed reality and robotics applications. Given the vast amount of available data nowadays, many applications constrain storage and/or bandwidth to work efficiently. To satisfy these constraints, many applications compress a scene representation by reducing its number of 3D points. While state-of-the-art methods use $K$-cover-based algorithms to compress a scene, they are slow and hard to tune. To enhance speed and facilitate parameter tuning, this work introduces a novel approach that compresses a scene representation by means of a constrained quadratic program (QP). Because this QP resembles a one-class support vector machine, we derive a variant of the sequential minimal optimization to solve it. Our approach uses the points corresponding to the support vectors as the subset of points to represent a scene. We also present an efficient initialization method that allows our method to converge quickly. Our experiments on publicly available datasets show that our approach compresses a scene representation quickly while delivering accurate pose estimates.
\end{abstract}
\vspace{-4mm}
\section{Introduction}
\label{sec:introduction}

\let\thefootnote\relax\footnote{This work was presented at IEEE 3DV 2020.}

Estimating the camera pose (\ie, position and orientation) is a crucial step for applications in self-driving cars~\cite{hane20173d,Lee_2013_CVPR,lee2013structureless}, robotics~\cite{davison2007monoslam,kaess2010probabilistic,kneip2013using}, and mixed reality~\cite{fragoso2011translatar,middelberg2014scalable,petter2011automatic,sweeney2015efficient,turk2015computer}. This is because these applications use camera poses to understand how the camera is positioned and oriented with respect to an environment.

While many 3D computer vision systems successfully localize themselves in an environment, they struggle to scale well when the environment becomes very large~\cite{sattler2017large}. Their struggle has various reasons. First, the memory and/or disk space requirements needed to store and represent the environment can be substantial. This is because the common scene representation can contain a collection of images, 3D points, and 2D image features with their respective feature descriptors (\eg, SIFT~\cite{lowe1999object}). Second, most of these systems use pose estimators that require longer time to operate when the representation of the scene is large. Although there exist efforts that increase efficiency of pose estimation  (\eg,~\cite{camposeco2018hybrid,camposeco2016minimal,sweeney2014gdls,sweeney2016large}), they still struggle when the scene representation is large.

\begin{figure}[t]
    \centering
    \includegraphics[width=\columnwidth]{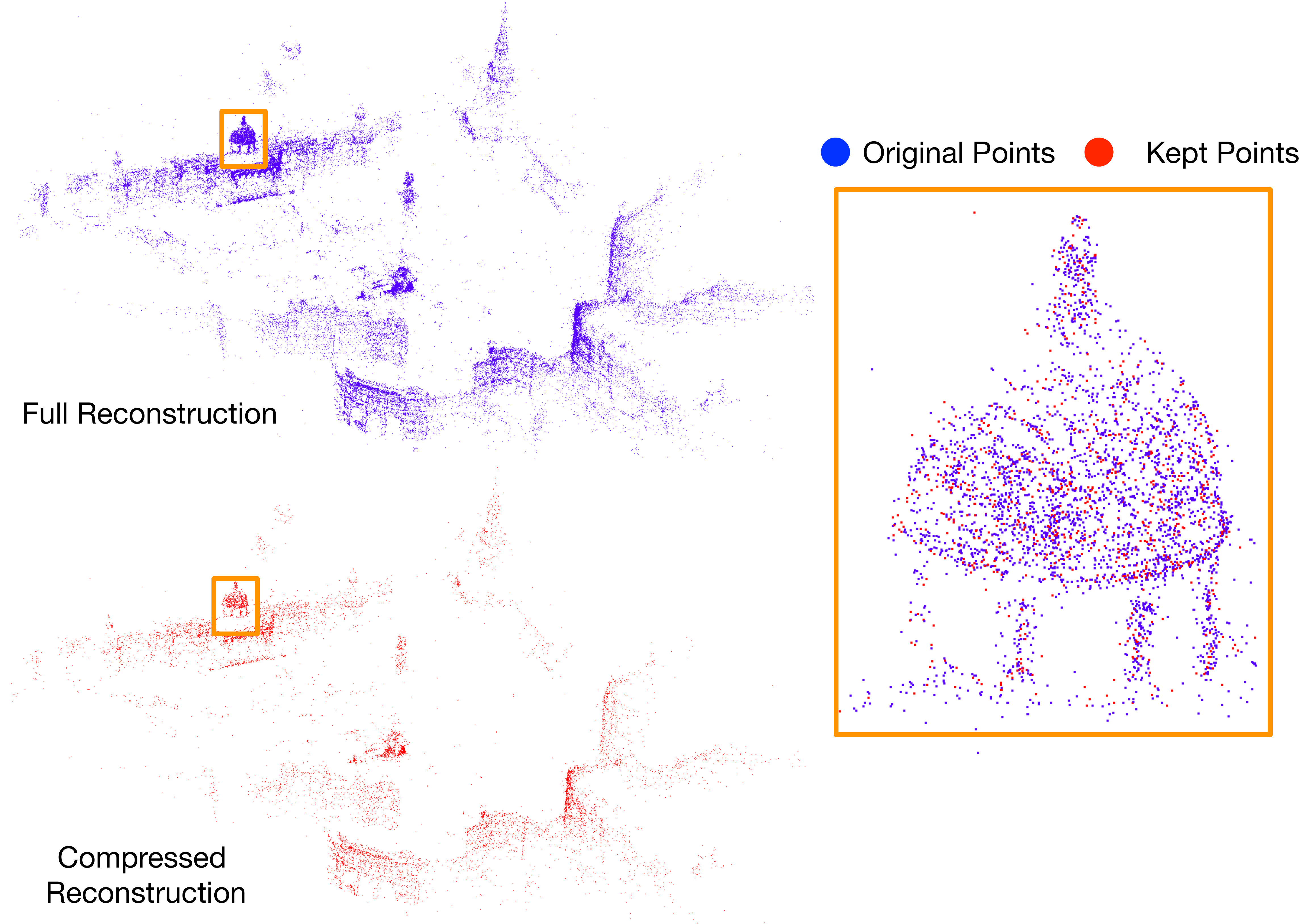}
    \vspace{-6mm}
    \caption{Unlike $K$-cover-based scene compression techniques, we formulate the scene representation compression problem with a quadratic program (QP). The goal of the QP is to keep 3D points that have good spatial coverage and visual distinctiveness. Inspired by one-class support vector machines (SVMs), we derive an efficient sequential-minimal-optimization (SMO) QP solver for scene compression. Similar to SVMs that use the support vectors to define its decision function, our method keeps the 3D points that are marked as the support vectors and discard the remaining ones.}
    \label{fig:teaser}
    \vspace{-4mm}
\end{figure}

To improve the scalability of these computer vision systems, we aim to compress scene representations. In particular, we focus on compressing point clouds computed via structure-from-motion (SfM) pipelines. This is because SfM point clouds are the most common scene representation for visual-based localization.

The reader must recall that an SfM point cloud has a collection of 3D points describing the geometry of a scene. Every point in this representation (typically) has a set of 2D image features and their respective feature descriptors~\cite{li2010location,sattler2011fast}. The goal of this work is to carefully select a subset of 3D points from an SfM point cloud such that the selection provides enough information to accurately estimate a camera pose. Consequently, compressing an SfM point cloud reduces the storage footprint because it prunes ``unnecessary'' points. This is useful especially for mobile agents (\eg, mobile devices, robots, etc.) that have limited storage and need to self-localize in an environment. 


Most state-of-the-art point cloud compression methods~\cite{camposeco2019hybrid,cao2014minimal,li2010location,soo20133d} use the $K$-cover-based methodology. In simple terms, the $K$-cover-based methodology aims to find a minimum subset of 3D points such that each database image sees at least $K$ points in the subset. While this approach effectively compresses an SfM point cloud, finding the right parameters to reduce the size of an SfM point cloud by a certain factor is not a trivial task.

We present an optimization problem for compressing an SfM point cloud that is convex and resembles the one-class support vector machine (SVM)~\cite{scholkopf2000support}; see Fig.~\ref{fig:teaser} for an illustration. Thanks to this resemblance, we derive an efficient solver based on the sequential minimal optimization (SMO)~\cite{platt1998sequential}, which is an efficient and scalable optimization technique to train SVMs. The proposed approach aims to select points that cover sufficiently the surface to represent but at the same time present a visual distinctiveness. From the SVM perspective, the support vectors~\cite{cortes1995support,scholkopf2002learning} correspond to the 3D-point selection that best represent a scene. Moreover, our proposed approach has parameters with intuitive meanings, which makes the parameter tuning simpler than that of $K$-cover-based methods. Our experiments on existing large-scale image-based localization datasets (\eg, Dubrovnik~\cite{sattler2011fast} and Cambridge Landmarks dataset~\cite{kendall2015posenet}) show that our approach compresses an SfM point cloud efficiently while yielding accurate pose estimates.

In sum, we present the following contributions: {\bf 1)} a QP convex formulation for compressing SfM point clouds with easy to tune parameters; {\bf 2)} an effective initialization method for the QP compression problem; and {\bf 3)} an efficient SMO-based constrained QP solver for compression.


\vspace{-1mm}
\section{Related Work}
\label{sec:rel_work}
\vspace{-1mm}

Reducing the number of points in an SfM point cloud has been addressed by means of $K$-cover-inspired algorithms, mixed-integer quadratic program (QP) optimization methods, and  deep-learning-based approaches. This section covers work that falls under these three different approaches.

\vspace{-1mm}
\subsection{$K$-cover-based Approaches}
\vspace{-1mm}
Li~\etal~\cite{li2010location} presented a $K$-cover-inspired algorithm to select a minimum subset of 3D points such that each database image sees at least $K$ points in the subset. While this approach works well for reducing an SfM point cloud size, computing the subset of 3D points is a challenging combinatorial problem. Li~\etal~\cite{li2010location} computes this minimum subset by incrementally building it. Their method uses a gain function that allows the algorithm to select points that contribute to the construction of the subset.

While the $K$-cover-inspired algorithm~\cite{li2010location} reduces an SfM point cloud size, it does not include information about the visual distinctiveness of each of the selected points. This aspect is important for image-based localization since visual features (\eg, SIFT~\cite{lowe1999object}) are crucial to establish 2D-to-3D correspondences which are the input for any pose estimator. To address this issue, Cao and Snavely~\cite{cao2014minimal} extended the $K$-cover algorithm~\cite{li2010location} by considering the coverage and the visual distinctiveness of the points. The coverage aspect imposes the constraint that the points in the subset are highly visible,~\ie, that a new camera observing the scene has a high probability of seeing most of the points in the subset. Moreover, their extension includes a visual distinctiveness term that favors the selection of points that are easy to visually identify. 

Camposeco~\etal~\cite{camposeco2019hybrid} presented a hybrid scene compression method that computes two sets of 3D points: the first set of points is small and contains raw descriptors while the second set is larger and contains quantized descriptors. The first set yields high-quality 2D-3D correspondences while the second set allows the hybrid compression method to verify hypotheses within a RANSAC loop. This hybrid method thus reduces the memory or storage footprint by decreasing the number of 3D points to keep and quantizing descriptors. At its core of this hybrid compression method is a variant of the $K$-cover algorithm and quantization methods all integrated with RANSAC. Although the $K$-cover-based methods effectively compress an SfM point cloud, it is hard to find the parameter $K$ such that the resultant compressed SfM point cloud reduces its size by a certain factor. In contrast to these methods, our approach has parameters that are easy to set given their intuitive meaning

\vspace{-1mm}
\subsection{Mixed-integer-QP-based Approaches}
\vspace{-1mm}
An alternative approach to the $K$-cover-based algorithms is a formulation using mixed-integer programming. Park~\etal~\cite{soo20133d} introduced a constrained quadratic program (QP) formulation mimicking the $K$-cover problem. This problem aims to compute a binary vector. The $i$-th entry of this vector is set to 1 when the $i$-th point is kept, and it is set 0 otherwise. Unfortunately, solving the formulated constrained QP is not scalable and requires specialized mixed-integer solvers. This method struggles to scale due to the $n \times n$ matrix that encodes pairwise relationships among the points; $n$ is the number of points. Clearly, for large-scale datasets $n$ is large and the scalability of this method depends of the used solver. Dymczyk~\etal~\cite{dymczyk2015keep} build on Park~\etal~\cite{soo20133d} work and scale it by dividing the problem into sub-problems. Although our proposed formulation also uses a constrained QP formulation, we present an efficient solver that scales well. This is because our QP formulation shares the structure of a one-class SVM~\cite{scholkopf2000support} and can be solved efficiently using a variant of the sequential minimal optimization~\cite{platt1998sequential} (SMO) method that we present in Sec.~\ref{sec:description}.

\vspace{-1mm}
\subsection{Deep-learning-based Approaches}
\vspace{-1mm}
Several deep-learning-based approaches~\cite{hochreiter1997long,kendall2017geometric,kendall2015posenet} aim to address image-based localization or pose estimation. These methods can be considered as compression approaches because the learned weights of the neural network encode the parameters of a scene. Kendall~\etal~presented PoseNet~\cite{kendall2015posenet}, a convolutional neural network (CNN) that estimates camera poses for relocalization. Walch~\etal~\cite{walch2017image} combined a CNN with LSTMs~\cite{hochreiter1997long} to estimate camera poses. While deep-learning-based approaches have shown impressive results, they still require specialized equipment (\eg, GPUs) to train them, struggle to generalize on unseen scenes, and are computationally expensive for mobile devices. In contrast, our proposed method does not require specialized equipment, has an explainable or interpretable compression model, and allows estimators to compute accurate poses quickly by using an SMO-based solver.
\section{Efficient Scene Compression}
\label{sec:description}

Our solution aims to select points that are visually distinct and far away from each other. Visually distinct points help feature matchers produce good correspondences, and consequently, produce good pose estimates. Having some distance between pairs of points enforces the solution to cover most of the scene and benefits the pose estimator. This is because 3D points that are far from each other often produce 2D-3D correspondences that impose constraints yielding good pose estimates. To find the set of 3D points that satisfy the aforementioned constraints, we present a novel convex optimization problem that can be solved efficiently. Unlike the state of the art which builds on the $K$-cover problem, the proposed formulation aims to learn a sparse discrete probability distribution over the set of points. This learned distribution has non-zero values on the selected points and zero on the points that are discarded.

Mathematically, we aim to learn a sparse distribution $\boldsymbol{\alpha}$ over the $m$ 3D points of the input point cloud. This sparse distribution has non-zero probabilities on those points that are considered visually distinct and have a reasonable average spatial distance with respect to their closest neighbors. To formulate this problem, we need to define terms that measure the spatial distance to their neighbors and visual distinctiveness of each of the points as a function of $\boldsymbol{\alpha}$. To measure the spatial distance among pairs of points, we use the following term:
\begin{equation}
    C = \boldsymbol{\alpha}^\intercal K \boldsymbol{\alpha},
    \label{eq:coverage}
\end{equation}
where the entries of the matrix $K \in \mathbb{R}^{m \times m}$ are
\begin{equation}
    K_{ij} = k(\mathbf{x}_i, \mathbf{x}_j) = \exp{\left(-\frac{\|\mathbf{x}_i - \mathbf{x}_j \|^2}{2 \sigma^2}\right)},
\end{equation}
$\mathbf{x}_i, \mathbf{x}_j \in \mathbb{R}^3$ are two point positions, and $\sigma$ is a parameter that controls when two points are considered close enough. Because the matrix $K$ is an RBF kernel~\cite{scholkopf2002learning}, the spatial distance term $C$ ranges between $0$ and $1$. It decreases when two points are far away and increases when two points are close to each other. To fulfill the goal of keeping points that are far apart of each other, we need to minimize $C$. 

The term $C$ can be interpreted as the expected spatial distance score among pairs of points, \ie, 
\begin{equation}
\mathbb{E}\left[k(\mathbf{x}_i, \mathbf{x}_j)\right] = \sum_{i,j} k(\mathbf{x}_i, \mathbf{x}_j) p(\mathbf{x}_i, \mathbf{x}_j) = \boldsymbol{\alpha}^\intercal K \boldsymbol{\alpha} = C
\end{equation}
where $\mathbb{E}\left[ \cdot \right]$ is the expectation operator, and $p(\mathbf{x}_i, \mathbf{x}_j) = p(\mathbf{x}_i)p(\mathbf{x}_j)=\alpha_i\alpha_j$ is the joint probability encoding the chances that the point pair $(\mathbf{x}_i, \mathbf{x}_j)$ is selected. By minimizing $C$ over $\boldsymbol{\alpha}$, we are indirectly maximizing the expected distance between the selected point pairs. Thus, by minimizing $C$ we enforce the algorithm to learn a distribution $\boldsymbol{\alpha}$ that aims to maximize the expected pairwise distance between points in the input point cloud. Since the algorithm learns a sparse distribution $\boldsymbol{\alpha}$, the algorithm thus learns to select only a handful of 3D points from the point cloud.

To keep the most visually distinctive or ``easy to match'' 3D points, we use the following term:
\begin{equation}
    D = \mathbf{d}^\intercal \boldsymbol{\alpha},
    \label{eq:distinctiveness}
\end{equation}
where $\mathbf{d} \in \mathbb{R}^m $ is a vector holding a visual distinctiveness score for every point in the input point cloud. Because our goal is to keep the most visually distinctive points, we need to maximize this term. Examples of visual distinctiveness scores are the average of the matching scores (\eg, descriptor distances) of every 3D point or the amount of 2D image features associated to a 3D point.

The term $D$ also can be interpreted as the expected visual distinctiveness score of the selected points, \ie, 
\begin{equation}
    \mathbb{E}\left[ d_i \right] = \sum_i d_i p(\mathbf{x}_i) = \sum_i d_i \alpha_i = \mathbf{d}^\intercal \boldsymbol{\alpha} = D,
\end{equation}
where $d_i$ is the visual distinctiveness score for the $i$-th point, and $\alpha_i = p(\mathbf{x}_i)$ is the probability of selecting the $i$-th point.

As stated above, we want to minimize $C$ (\ie, maximize the expected spatial distance among pairs of points) and maximize $D$ (\ie, keep the most visually distinctive points). To this end, we use the following cost function:
\begin{equation}
    J = C - \tau D = \boldsymbol{\alpha}^\intercal K \boldsymbol{\alpha} - \tau \mathbf{d}^\intercal \boldsymbol{\alpha},
    \label{eq:cost_function}
\end{equation}
where $\tau$ is a scalar that controls the trade-off between spatial distance term $C$ and the visual distinctiveness $D$. By minimizing $J$ over $\boldsymbol{\alpha}$, we minimize $C$ and maximize $D$. When visual distinctiveness is more important, then $\tau$ must be high. On the other hand, when the spatial distance term is more important, then $\tau$ must be near zero.

To learn the sparse distribution $\boldsymbol{\alpha}$ that minimizes $J$, we solve the following optimization problem:
\begin{equation}
    \begin{aligned}
& \underset{\boldsymbol{\alpha}}{\text{minimize}}
& & \boldsymbol{\alpha}^\intercal\mathbf{K}\boldsymbol{\alpha} - \tau \mathbf{d}^\intercal\boldsymbol{\alpha}\\
& \text{subject to}
& & \sum_i ^m \alpha_i =1,\\
& && \quad 0\leq\alpha_i\leq \frac{1}{\nu m};\quad i = 1, \ldots, m,
\end{aligned}
\label{eq:qp_problem}
\end{equation}
where $\nu \in \left(0, 1\right]$ is the compression factor, a scalar that controls the sparsity of the distribution $\boldsymbol{\alpha}$. This parameter thus controls the compression rate. This is because when $\nu=1$, then $\boldsymbol{\alpha}$ becomes a uniform distribution, which is equivalent to no compression. On the other hand, when $\nu < 1$, then we allow the algorithm to put more mass on a few points. In this case, this is equivalent to select only a few points which reduces the size of a point cloud. 

The problem shown in Eq.~\eqref{eq:qp_problem} is a quadratic program (QP). As such, our compression problem is convex. This is because the RBF kernel matrix $K$ is positive-semi-definite matrix~\cite{scholkopf2002learning}, and the problem has linear equality and inequality constraints. Any convex solver (\eg, Newton-like solvers~\cite{boyd2004convex}) can be used to find $\boldsymbol{\alpha}$. Unfortunately, these methods do not scale well when the number of unknown variables is large. Nevertheless, we can exploit the intimate relationship that this QP problem has with one-class SVMs to derive an efficient solver.

\subsection{Relation with One-class SVMs}
The proposed problem in Eq.~\eqref{eq:qp_problem} has a direct relationship to one-class classifiers. We can obtain the exact one-class SVM dual formulation~\cite{scholkopf2000support} by setting $\tau=0$. With this setting, we omit the linear term $D$ and only keep the spatial distance term $C$. This reveals that the proposed approach with this setting reduces a point cloud by keeping the support vectors. Recall that the support vectors have a corresponding non-zero entry in $\boldsymbol{\alpha}$, and are the points that allow an SVM to define a decision boundary for recognition. 

While this relationship provides insight as to how the proposed algorithm operates, it also enables an opportunity to derive an efficient solver. This is because efficient and scalable solvers that train an SVM exist, \eg, the sequential-minimal optimization (SMO)~\cite{platt1998sequential}. Unfortunately, we cannot directly use the one-class-SVM SMO solver for the proposed problem. There are two reasons that limit the SMO solver for our proposed problem. The first one is that the original one-class-SVM SMO solver does not consider a linear term; in our case the distinctiveness term $D$. The second reason is that the SMO uses the SVM decision rule to determine efficient variable updates. Our proposed problem is not a classification one. As such, our problem does not have a decision rule. Nevertheless, as we discuss in the next section, it is still possible to derive an SMO-like solver that can efficiently solve the proposed problem.

\subsection{Efficient SMO-like Solver}

Inspired by the SMO solver, we aim to formulate the simplest sub-problem that we can sequentially solve at a time. By solving these sub-problems sequentially, we can find the solution for the proposed problem. As shown by Platt~\cite{platt1998sequential}, the simplest problem that we can solve in an SVM involves a quadratic program with only two variables. The advantage of solving a QP problem with two variables is that we can solve it analytically. Consequently, we avoid expensive matrix operations (\eg, matrix inversions and multiplications) which are fundamental operations in Newton-based optimization methods.

To obtain the simplest QP problem with two variables, we need to algebraically manipulate Eq.~\eqref{eq:cost_function}. Recall that the goal is to obtain a cost function $J^\prime$ that focuses on only two variables: $\alpha_i$ and $\alpha_j$ which are the $i$-th and $j$-th entries of $\boldsymbol{\alpha}$, respectively, and $i \neq j$. After algebraic manipulations, we obtain the following $J^\prime$:
\begin{equation}
\begin{split}
J^\prime(i,j) &= \alpha_i^2 + 2 \alpha_i \alpha_j K_{ij} + \alpha_j^2 \\
&+ 2 \alpha_i \sum_{l \neq i, l \neq j} \alpha_j K_{il} + 2 \alpha_j  \sum_{l \neq i, l \neq j} \alpha_l K_{jl} \\ 
&- \tau d_i \alpha_i  - \tau d_j \alpha_j + g\left(\left\{ \alpha_t : t \neq i, j\right\}\right),
\end{split}
\end{equation}
where $g(\cdot)$ is a function including all the remaining entries $\left\{ \alpha_t : t \neq i, j\right\}$ in $\boldsymbol{\alpha}$. For the full derivation of $J^\prime$, we refer the reader to the appendix.

The original problem shown in Eq.~\eqref{eq:qp_problem} aims to learn a probability distribution $\boldsymbol{\alpha}$. Since $J^\prime$ focuses on only two variables, we need to update the equality constraints when we only optimize for the two entries $\alpha_i$ and $\alpha_j$. To do this, we need to ensure that the sum of all the entries in $\boldsymbol{\alpha}$ equals to one. At the same time we still need to ensure the inequality constraints shown in Eq.~\eqref{eq:qp_problem}. After considering these aspects, we obtain the following the problem:
\begin{equation}
\begin{aligned}
& \underset{\alpha_i, \alpha_j}{\text{minimize}}
& & J^\prime(i, j)\\
& \text{subject to}
& & \alpha_i + \alpha_j = \Delta,\\
& & & \quad 0\leq\alpha_i, \alpha_j\leq \frac{1}{\nu m},
\end{aligned}
\label{eq:simplest_cost}
\end{equation}
where $\Delta$ is the joint probability mass between $\alpha_i$ and $\alpha_j$. This means that we can minimize $J^\prime$ as long as we maintain the probability mass $\Delta$ between $\alpha_i$ and $\alpha_j$ constant. Similar to the SMO, this constraint imposes a solution over a line $\alpha_i + \alpha_j = \Delta$ and keeps a valid sum: $\sum_i \alpha_i = 1$. This is because the solver assumes that the starting probability distribution $\boldsymbol{\alpha}$ is feasible, \ie, it sums up to one and satisfies the inequality constraints.

Similar to the SMO algorithm, the proposed algorithm needs to iteratively select a pair of variables $\alpha_i$ and $\alpha_j$, and solve the problem shown in Eq.~\eqref{eq:simplest_cost}. To solve this simplified problem analytically, we leverage the equality constraint to set $\alpha_j = \Delta - \alpha_i$. Via substitution, we simplify the cost function $J^\prime(\alpha_i)$:
\begin{equation}
\begin{split}
J^\prime(\alpha_i) &= \alpha_i^2 + 2\alpha_1\left(\Delta - \alpha_i\right) K_{ij} + (\Delta - \alpha_i)^2 \\
 &+ 2\alpha_i \sum_{l \neq i, l \neq j} \alpha_l K_{il} \\
 &+ 2 \left( \Delta - \alpha_i\right) \sum_{l \neq i, l \neq j} \alpha_l K_{jl} \\
 &- \tau \left( d_i \alpha_i + d_j\left(\Delta - \alpha_i \right) \right) + g\left( \left\{ \alpha_t : t \neq i, j\right\} \right).
\end{split}
\end{equation}

Given that $J^\prime(\alpha_i)$ is a function of a single variable, we can obtain the optimal $\alpha^\star_i$ analytically by solving $\frac{\partial J^\prime}{\partial \alpha_i} = 0$. This analytical solution is:
\begin{equation}
    \alpha_i^\dag = \frac{1}{2} \left( \frac{T}{2 \left(1 - K_{ij}\right) + \Delta }\right),
    \label{eq:unboxed_soln}
\end{equation}
where 
\begin{equation}
    T = \tau (d_i - d_j) - 2 \sum_{l \neq i, l \neq j} \alpha_l K_{il} +  2 \sum_{l \neq i, l \neq j} \alpha_l K_{jl}.
\end{equation}

The solution $\alpha_i^\dag$ does not consider the inequality constraints shown in Eq.~\ref{eq:qp_problem}. To enforce these inequality constraints, we box the solution~$\alpha_i^\dag$, \ie, 
\begin{equation}
    \alpha_i^\star = \max \left(0, \min\left( \min\left( \frac{1}{\nu m}, \Delta \right), \alpha^\dag \right)\right).
    \label{eq:boxed_solution}
\end{equation}
The $\min(\cdot)$ operations above ensures that the upper bounds are satisfied. On the other hand, the $\max(\cdot)$ operation makes sure the lower bound is enforced. Finally, we compute $\alpha_j = \Delta - \alpha_i^\star$.


\subsection{Algorithm}
\label{sec:algorithm}

\begin{algorithm}[t]
\footnotesize{
    \SetAlgoNoLine
    \SetKwInOut{Input}{Input}
    \SetKwInOut{Output}{Output}
    \SetKwInOut{Params}{Parameters}
    \Input{Set of $m$ 3D points and their corresponding distinctiveness scores $\mathcal{M} = \{ \left(\mathbf{x}_i, d_i\right) \}_{i=1}^m$.}
    \Output{Probability distribution $\boldsymbol{\alpha} \in \mathbb{R}^m$.}
    \Params{Compression factor $\nu$ \\ Trade-off factor $\tau$ \\ RBF kernel bandwidth $\sigma$}

   \text{// Initialize with a feasible probability distribution.} \\
   $\boldsymbol{\alpha}$ = InitializeProbability($\mathcal{M}, \nu$) \\
   \Repeat{Convergence}
   {
      $i, j = \text{SelectPair}(m)$ \\
      $\Delta = \alpha_i + \alpha_j$ \\
      \text{// Compute Eq.~\eqref{eq:boxed_solution}.} \\
      $\alpha_i^\star = \text{ComputeProbability}(i, j, \Delta, \nu, m, \sigma, \mathcal{M})$ \\
      \text{// Update probability pair.} \\
      $\alpha_i = \alpha_i^\star$ \\
      $\alpha_j = \Delta - \alpha_i$
   }
}
   \caption{Efficient Scene Compression}
   \label{alg:compression}
\end{algorithm}

To find the probability distribution $\boldsymbol{\alpha}$ and identify the 3D points to keep, we need to solve the QP problem shown in Eq.~\eqref{eq:qp_problem}. The solution of this QP problem requires selecting a sequence of pairs of points and solving a simpler QP problem for each pair (see Eq.~\eqref{eq:simplest_cost}). The solution of each of the simpler QP problems can be computed via Eq.~\eqref{eq:unboxed_soln} and Eq.~\eqref{eq:boxed_solution}. Our proposed SMO-solver is guaranteed to converge as it can be seen as a special case of the generalized SMO algorithm~\cite{keerthi2002convergence}. Algorithm~\ref{alg:compression} summarizes the SMO-based point-cloud compression procedure.

{\noindent \bf Initialization.} Given the set of $m$ 3D points and their corresponding visual distinctiveness scores $\mathcal{M} = \left\{ \left( \mathbf{x}_i, d_i \right) \right\}_{i=1}^m$, the efficient scene compression algorithm first initializes $\boldsymbol{\alpha}$ with a feasible probability distribution satisfying the inequality constraints from Eq.~\eqref{eq:qp_problem} (see step 2 in Algorithm~\ref{alg:compression}). To satisfy the inequality constraints, the algorithm first sets $\boldsymbol{\alpha} = \mathbf{0}$. Subsequently, it ranks the $m$ 3D points based on the visual distinctiveness score vector $\mathbf{d}$. Then, the algorithm selects the $n = \ceil{\nu m}$ most visually distinctive points by using their corresponding scores in $\mathbf{d}$, and sets their corresponding probability entries in $\boldsymbol{\alpha}$ to $\frac{1}{\nu m}$. When the sum of the probabilities exceeds $1$, then the algorithm identifies one of the selected $n$ points and updates its probability entry such that the sum of the entries in $\boldsymbol{\alpha}$ equals $1$. Using the $n$ most visually distinctive points to initialize $\boldsymbol{\alpha}$ helps the algorithm search for the solution from a set of points that are likely to match more easily and quickly. This initialization method is inspired on SVM initialization methods~\cite{platt2001estimating}.


{\noindent \bf Pair selection.} Given the initialized $\boldsymbol{\alpha}$, the algorithm starts solving the sequence of simpler QP problems (steps $2$ - $11$). To construct a simple QP problem shown in Eq.~\eqref{eq:simplest_cost}, the algorithm first selects a pair of points $(i, j)$. The simplest selection process is a random pick of two different points. A more sophisticated process selects a pair of points such that at least one of their corresponding probabilities is non-zero. This sophisticated process likely will decrease the number of iterations since a pair with zero probabilities does not produce an update on $\boldsymbol{\alpha}$. However, this process requires tracking the non-zero probabilities which increases the computational complexity.

{\noindent \bf Updating pair probabilities.} Given a pair $(i, j)$, the algorithm first computes the probability mass $\Delta$ that needs to be redistributed (step 5). Subsequently, the algorithm computes the probability $\alpha_i^\star$ for the $i$-th point using Eq.~\eqref{eq:boxed_solution} and assigns it to $\alpha_i$ in step 9. Finally, the algorithm computes the probability $\alpha_j$ for the $j$-th point in step 10.

Similar to modern SVM solvers~\cite{chang2011libsvm,joachims1998making}, the implementation of the efficient scene compression summarized in Algorithm~\ref{alg:compression} uses a cache to alleviate the storage and computational cost of the RBF kernel function. Using this cache avoids storing the kernel matrix $K$ which can become prohibitively expensive as it requires $\mathcal{O}(m^2)$ storage cost.

The theoretical stopping criterion of the algorithm is based on the Karush-Kuhn-Tucker (KKT)~\cite{boyd2004convex} conditions of a QP. Unfortunately, evaluating the KKT conditions given a large set of $m$ points also requires a large storage footprint. Consequently, using this theoretical stopping criterion becomes impractical. Unlike the supervised classification or regression SVM problems that have stopping criteria based on the expected targets, the scene compression problem lacks any of these expected targets. Consequently, the stopping criteria from the publicly available SVMs do not apply for this problem. Inspired by recent large-scale optimization techniques used in deep learning~\cite{bottou2010large,goodfellow2016deep,adam2015}, the implementation of the optimal scene compression procedure shown in Algorithm~\ref{alg:compression} runs the loop in steps 3 to 11 for a fixed number of iterations. 
\vspace{-4mm}
\section{Experiments}
\label{sec:experiments}
\vspace{-2mm}

The goals of the experiments in this section are to measure 1) the compression times; 2) storage footprint of a compressed scene; 3) the localization rate; and 4) the accuracy of the estimated poses using the compressed scenes.

{\noindent \bf Datasets.} We used Kings College, Old Hospital, Shop Facade, and St. Mary's Church datasets from the Cambridge Landmark dataset collection~\cite{kendall2015posenet}, and Dubrovnik~\cite{li2010location}. These datasets present different imaging conditions, a wide range of number of images, and large number of 3D points in the reconstructions. The Cambridge Landmarks datasets~\cite{kendall2015posenet} contain a set of query images for which visual-based localization systems need to estimate their camera poses, a reference SfM reconstruction, and their corresponding set of images and features. For the Dubrovnik~\cite{li2010location} dataset, we used the protocol presented by Li~\etal~\cite{li2010location} where a few images and points are removed from the full reconstructions and then used as query images. To compute the localization accuracy, we compare the estimated poses and those of the original reconstruction. 

{\noindent \bf Implementation Details.} We implemented our compression procedure\footnote{Code:\url{https://github.com/mameratrujillo/Efficient_Scene_Compression}} and an image-based localization system in C++ using the Theia SfM library~\cite{sweeney2015theia}. We used Root-SIFT~\cite{arandjelovic2012three} descriptors, FLANN~\cite{muja2014scalable} as an approximate nearest-neighbor matcher, and P3P~\cite{kneip2011novel} as the camera pose estimator. We build a FLANN index for every dataset to better represent their Root-SIFT distribution~\cite{cui2017graphmatch}. We ran our algorithm for up to 4096 iterations because our initialization process returns a good feasible solution. We used a radial-basis-function (RBF) to map low average descriptor distances to higher values. This is needed because our approach assumes that higher visual distinctiveness scores correspond to more distinctive features.

{\noindent \bf Visual distinctiveness scores. } The experiments use three different scores: {\bf 1)} the average descriptor distance from the pair-wise image matches coming from the SfM tracks; we use an exponential mapping to assign a high score for lower descriptor distance and a higher one for a larger distance; {\bf 2)} the fraction of cameras that see a 3D point normalized by the total number of cameras; and {\bf 3)} the fraction of cameras that see a 3D point normalized by the maximum number of cameras observing a 3D point in the input reconstruction. See the appendix material for more details on these scores.

\vspace{-2mm}
\subsection{Ablation Studies}
\vspace{-1mm}


The experiments in this section show the effect on localization performance of the compression factor $\nu$, spatial-distance-and-visual-distinctiveness trade-off parameter $\tau$, and the RBF kernel bandwidth $\sigma$. We use the average descriptor distance visual distinctiveness score.


\begin{figure}[t]
    \centering
    \includegraphics[width=\columnwidth]{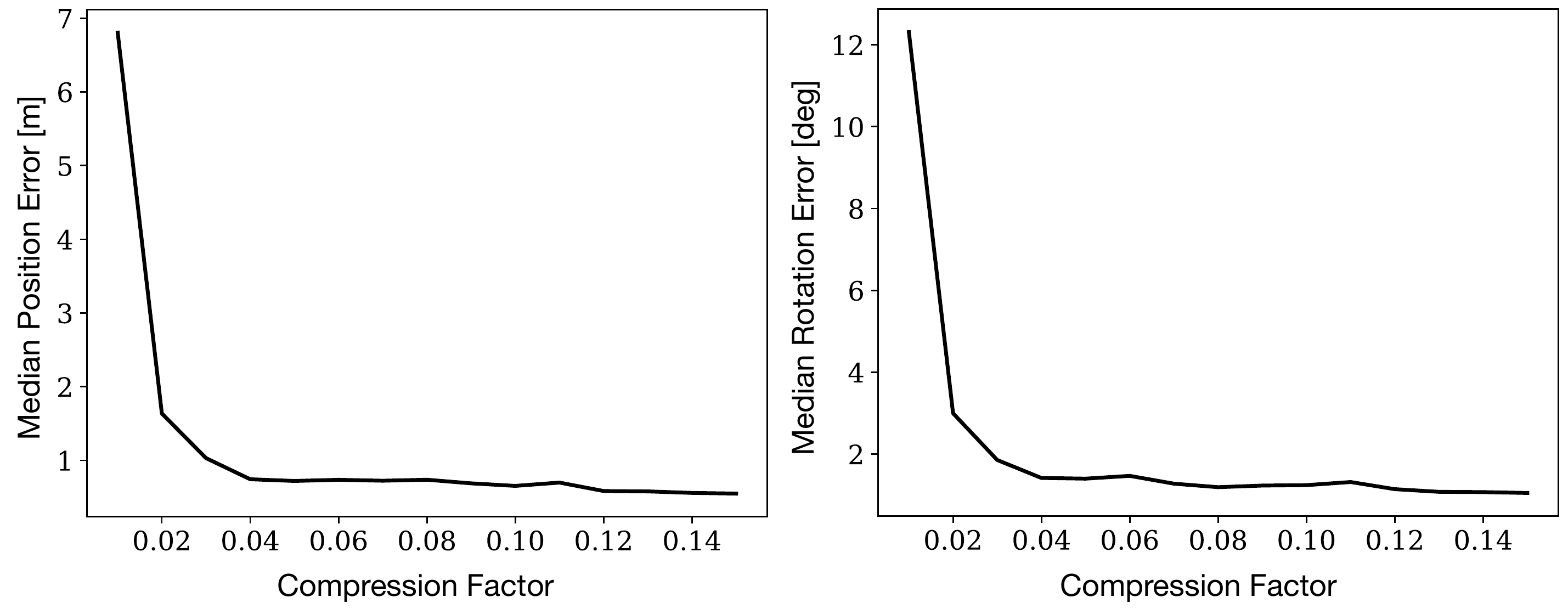}
    \vspace{-6mm}
    \caption{Pose errors as a function of compression factor $\nu$. The larger the compression factor $\nu$, the smaller the position and rotation errors (left and right, respectively).}
    \label{fig:cf_ablation}
    \vspace{-5mm}
\end{figure}

{\noindent \bf The compression factor impacts the localization performance.}~We measured the position and rotation errors of a localized camera as a function of the compression factor $\nu$. For this experiment, we used the Shop Facade~\cite{kendall2015posenet} dataset. In Fig.~\ref{fig:cf_ablation} we can see that lower compression factors $\nu < 0.05$ return higher errors. This is because the compressed scene contains fewer 3D points and this makes the image-based localizer struggle to operate optimally. However, the errors reduce considerably when we use $\nu > 0.05$.

\begin{figure*}[t]
    \centering
    \includegraphics[width=\textwidth]{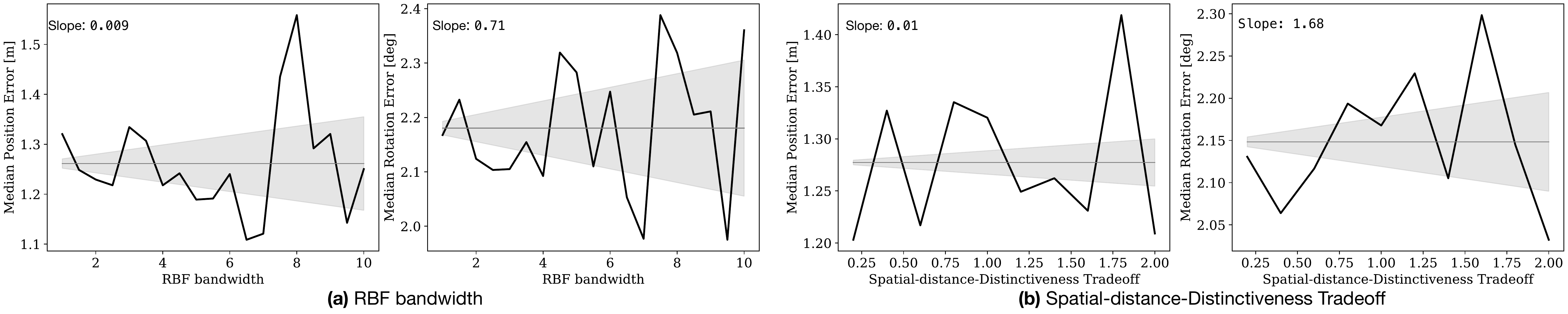}
    \vspace{-5mm}
    \caption{{\bf (a)} Median position and rotation errors as a function of RBF kernel bandwidth $\sigma$ and {\bf (b)} spatial-distance-visual-distinctiveness trade-off $\tau$. The errors vary around a common value (horizontal gray line) due to RANSAC, and observe that the variation increases as a function of $\tau$ and $\sigma$ (shaded gray area). The slopes of the lines enclosing the variation are larger for those of $\tau$ than that of $\sigma$, especially for rotation errors. Therefore, $\tau$ induces a larger uncertainty and thus impacting more the localization performance than that of $\sigma$.}
    \label{fig:cdt_rbf_ablation}
    \vspace{-5mm}
\end{figure*}

\begin{figure}[t]
    \centering
    \includegraphics[width=0.90 \columnwidth]{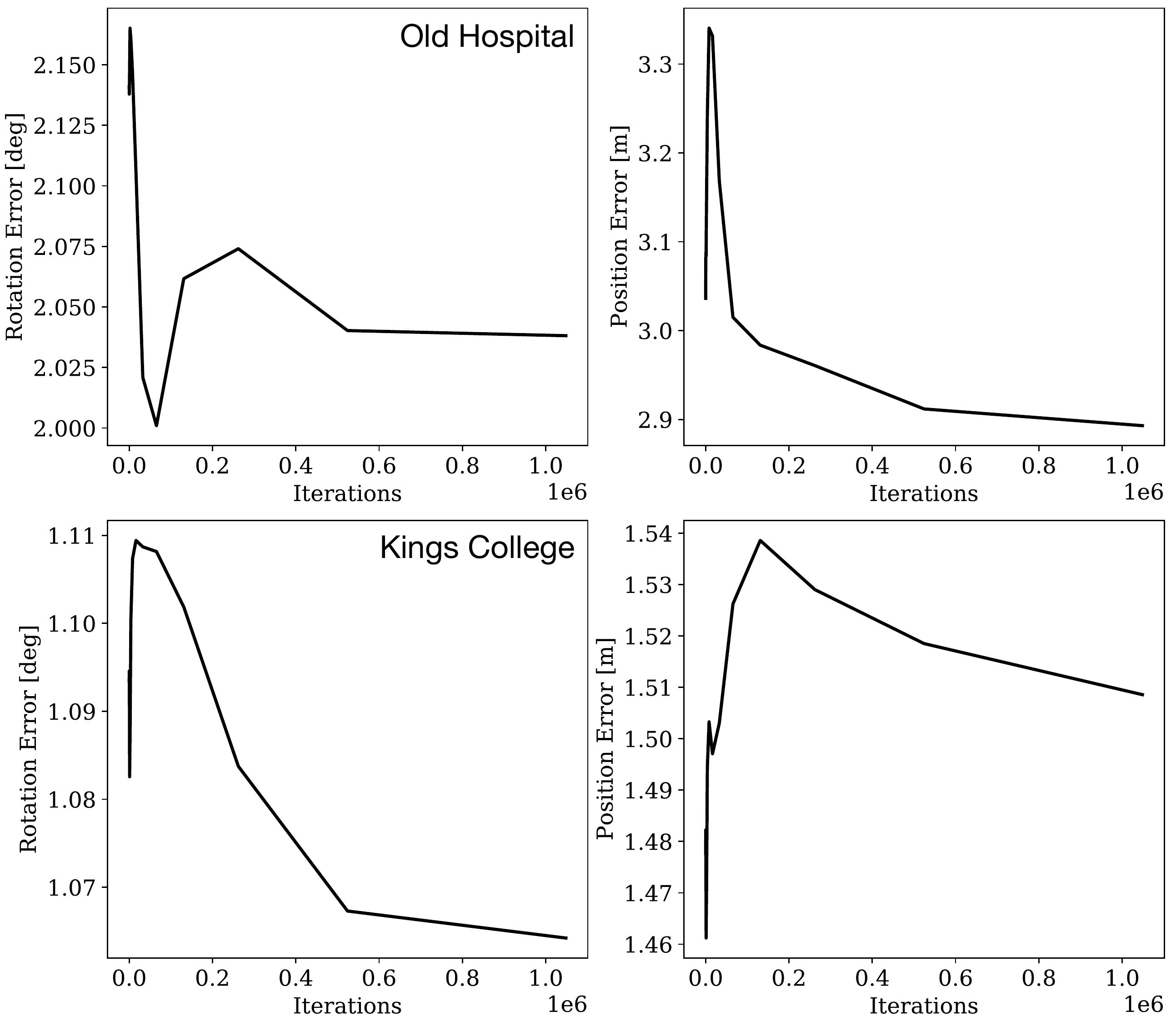}
    \vspace{-1mm}
    \caption{Rotation and position errors as a function of the number of iterations on Old Hospital (top row) and King's Collegue (bottom row). The initialization method gives a good starting solution and the optimization modestly improves pose accuracy.}
    \label{fig:acc_vs_iterations}
    \vspace{-4mm}
\end{figure}

{\noindent \bf The spatial-distance-and-visual-distinctiveness trade-off $\tau$ tends to impact more the accuracy than the RBF kernel bandwidth $\sigma$.}~For this experiment we set $\nu=0.05$ and use the Shop Facade~\cite{kendall2015posenet} dataset. We vary $\tau$ 
and $\sigma$ in the ranges $\left[0, 2\right]$ and $\left[1, 10\right]$, respectively. We set $\sigma=1$ when we vary $\tau$, and set $\tau=1$ when we vary $\sigma$. Fig.~\ref{fig:cdt_rbf_ablation}(a) and Fig.~\ref{fig:cdt_rbf_ablation}(b) show the localization performance as a function of $\sigma$ and $\tau$, respectively. Because this experiment computes position and rotation errors as a function of RANSAC estimates, we observe that the median position and rotation errors tend to vary around a common value for both $\sigma$ and $\tau$ (horizontal gray line). However, we observe that the std. deviations of the errors (gray shaded area) as a function of $\sigma$ and $\tau$ tend to increase as we increase $\sigma$ and $\tau$. We fit a line to these std. deviations and compare their slopes to measure their rate of growth (shown in the top left of each plot). The slopes of the position errors are comparable for both $\sigma$ and $\tau$. However, the slope of rotation error when varying $\tau$ is greater than that of $\sigma$ by about $2\times$. This suggests that $\tau$ affects more rotation performance than $\sigma$.

\vspace{-2mm}
\subsection{Compression Times and Registration Accuracy}
\vspace{-2mm}
\begin{figure}[t]
    \centering
    \includegraphics[width=\columnwidth]{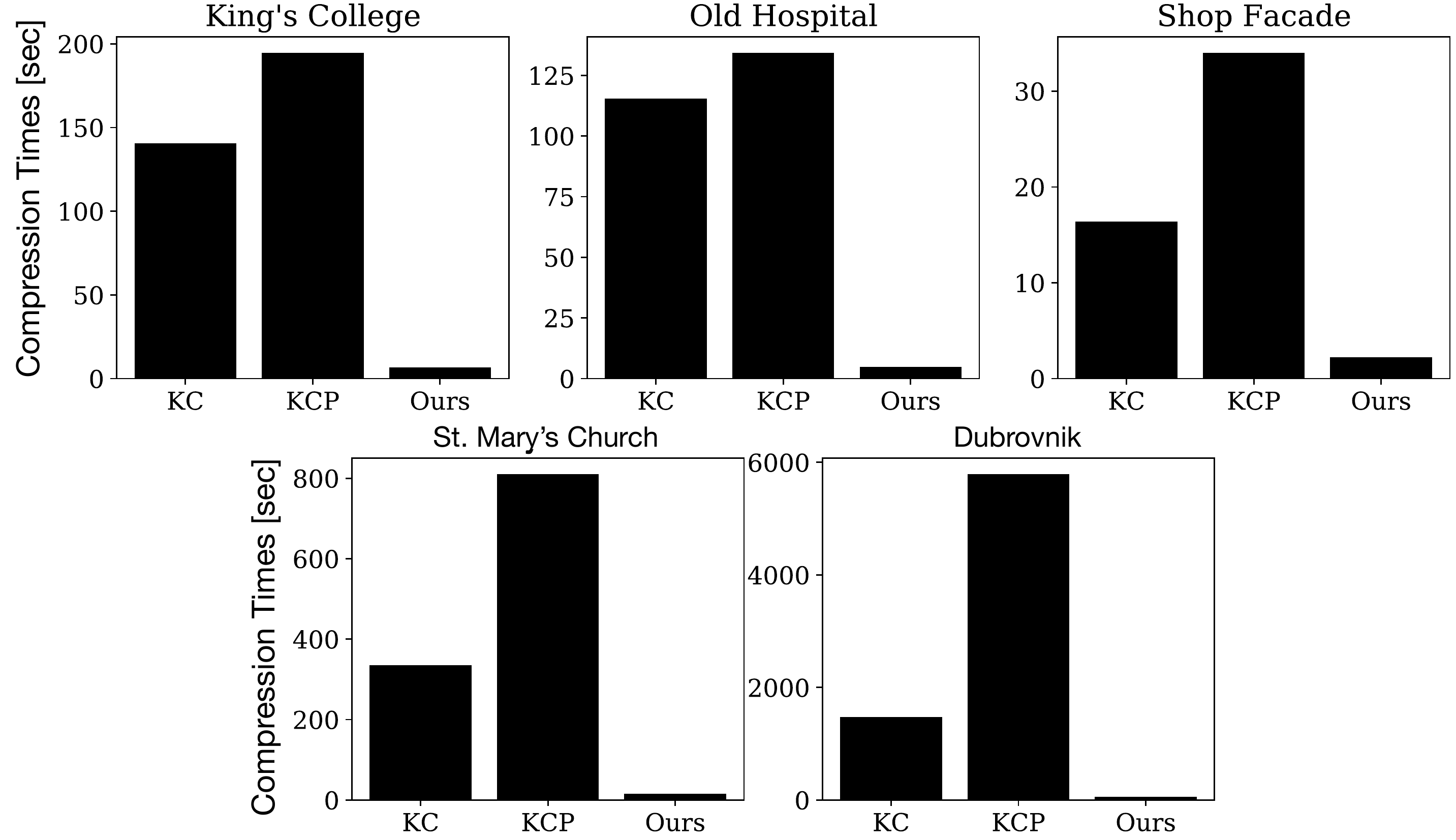}
    \vspace{-5mm}
    \caption{Compression times in seconds of KC~\cite{li2010location}, KCP~\cite{cao2014minimal}, and our approach on Cambridge Landmarks dataset~\cite{kendall2015posenet}. KC and KCP require a significant amount of time compared to our approach.}
    \label{fig:timing_results}
    \vspace{-5mm}
\end{figure}

\begin{table*}[t]
    \centering
    \caption{Localization performance and compression results on the Cambridge Landmarks datasets~\cite{kendall2015posenet}. The Table presents the size in MBs of the compressed reconstructions (MB Used columns) and position and orientation errors (Errors columns). The results of the bottom 9 rows come from Camposeco~\etal~\cite{camposeco2019hybrid} while the results of the top seven rows were computed using Theia library. Our method using different visual distinctiveness scores (shown in bold) achieves a comparable localization accuracy.}
    \label{tab:posenet_datasets}
    \footnotesize{
    \setlength{\tabcolsep}{1.8pt} 
    \begin{tabular}{l >{\centering}p{0.8cm} >{\centering}p{1.4cm} >{\centering}p{0.8cm} >{\centering}p{1.4cm} >{\centering}p{0.8cm} >{\centering}p{1.4cm} >{\centering}p{0.8cm} p{1.4cm}}
    \toprule
    \multirow{2}{*}{\bf Method} & \multicolumn{2}{c}{\bf King’s College}& \multicolumn{2}{c}{\bf Old Hospital} & \multicolumn{2}{c}{\bf Shop Facade} & \multicolumn{2}{c}{\bf St. Mary's Church} \\ 
     & MB Used & Errors $\left[m, \circ\right]$ & MB Used & Errors $\left[m, \circ\right]$ & MB Used & Errors $\left[m, \circ\right]$ & MB Used &  Errors $\left[m, \circ\right]$ \\
     \midrule
     {\bf Avg. descriptor distance (initial soln.)} & 2.2 & 1.48, 1.08 & 1.1 & 2.97, 2.14 & 0.41 & 0.75, 1.48 & 3.3 & 0.60, 0.89\\
     {\bf Avg. descriptor distance} & 2.2 & 1.53, 1.09 & 1.1 & 0.90, 2.17 & 0.41 & 0.72, 1.40 & 3.3 & 0.56, 0.89\\
     {\bf Num. cameras per point} & 2.2 & 0.94, 0.57 & 1.1 & 1.33, 0.99 & 0.41 & 0.31, 0.56 & 3.3 & 0.43, 0.64\\
     {\bf Max. num. cameras per point} & 2.2 & 0.92, 0.59 & 1.1 & 1.24, 0.96 & 0.41 & 0.31, 0.59 & 3.3 & 0.46, 0.65 \\
     KC~\cite{li2010location} & 3.1 & 1.48, 1.23 & 6.0 & 1.35, 1.06 & 0.85 & 0.51, 0.87 & 18 & 0.46, 0.69\\
     KCP~\cite{cao2014minimal} & 5.9 & 0.99, 0.86 & 8.2 & 1.19, 1.00 & 1.3 & 0.44, 0.80 & 24 & 0.40, 0.61\\
     No compression & 98 & 0.57, 0.50 & 34 & 0.96, 0.79 & 11 & 0.23, 0.43 & 131 & 0.28, 0.45\\
     \midrule
     Hybrid comp.~\cite{camposeco2019hybrid} & 1.01 & 0.81, 0.59 & 0.62 & 0.75, 1.01 & 0.16 & 0.19, 0.54 & 1.34 & 0.50, 0.49 \\
     DenseVLAD~\cite{torii201524} & 10.06 & 2.80, 5.72 & 13.98 & 4.01, 7.13 & 3.61 & 1.11, 7.61 & 23.23 & 2.31, 8.00\\
     PoseNet~\cite{kendall2015posenet} & 50 & 1.92, 5.40 & 50 & 2.31, 5.38 & 50 & 1.46, 8.08 & 50 &  2.65, 8.48\\
     Bayes PoseNet~\cite{kendall2016modelling} & 50 & 1.74, 4.06 & 50 & 2.57, 5.14 & 50 & 1.25, 7.54 & 50 & 2.11, 8.38\\
     LSTM PoseNet~\cite{walch2017image} & $\approx$ 50 & 0.99, 3.65 & $\approx$ 50 & 1.51, 4.29 & $\approx$ 50 & 1.18, 7.44 & $\approx$ 50 &  1.52, 6.68\\
     $\sigma^2$ PoseNet~\cite{kendall2017geometric} & $\approx$ 50 & 0.99, 1.06 & $\approx$ 50 & 2.17, 2.94 & $\approx$ 50 & 1.05, 3.97 & $\approx$ 50 &1.49, 3.43\\
     Geom. PoseNet~\cite{kendall2017geometric} & $\approx$ 50 & 0.88, 1.04 & $\approx$ 50 & 3.20, 3.29 & $\approx$ 50 & 0.88, 3.78 & $\approx$ 50 & 1.57, 3.32\\
     DSAC++~\cite{brachmann2018learning} & 207 & 0.18, 0.30 & 207 & 0.20, 0.30 & 207 & 0.06, 0.30 & 207 & 0.13, 0.40\\
     Active Search~\cite{sattler2016efficient} & 275 & 0.57, 0.70 & 140 & 0.52, 1.12 & 38.7 & 0.12, 0.41 & 359 & 0.22, 0.62\\
     \bottomrule
    \end{tabular}
    }
    \vspace{-5mm}
\end{table*}

The goals of this experiment are twofold: 1) measure the time a compression algorithm takes to reduce the size of an SfM point cloud; and 2) evaluate the localization performance using these compressed scenes. 

{\noindent \bf Our method is on average 20x faster than the k-cover-based methods.} We used the k-cover (KC)~\cite{li2010location} and probabilistic k-cover (KCP)~\cite{cao2014minimal} methods as baselines. We used publicly available implementations of these methods\footnote{\url{https://github.com/caosong/minimal_scene}}. The experiment did not include the hybrid-compression method by Camposeco~\etal~\cite{camposeco2019hybrid} since there is not a publicly available implementation. We ran these experiments on a machine with 32GB of RAM and an Intel i7 with 6 cores. Fig.~\ref{fig:timing_results} shows the compression times in seconds. This speed-up is due to the good initialization point described in Sec.~\ref{sec:algorithm} which mainly computes a feasible initial point using the average descriptor distance visual distinctiveness score. The time our method requires depends linearly on the number of iterations as the Sec.~\ref{sec:algorithm} states.

\begin{table}[t]
    \centering
    \caption{Localization and compression performance on the Dubrovnik~\cite{li2010location} dataset. The Table presents the same format as in Table~\ref{tab:posenet_datasets}. Our approach achieves comparable localization accuracy.}
    \footnotesize{
    \begin{tabular}{lc c c c}
    \toprule
     & {\bf Ours} & KC~\cite{li2010location} & KCP~\cite{cao2014minimal} & No comp.\\
    \midrule
    MB Used & 22 & 51 & 65 & 318 \\
    Errors  $\left[m, \circ\right]$ & 1.79, 0.56 & 1.99, 0.60 & 1.82, 0.55 & 1.31, 0.44 \\
    \bottomrule
    \end{tabular}
    }
    \label{tab:dubrovnik_aachen}
    \vspace{-6mm}
\end{table}

To measure the localization performance, we compute the rotation and position errors \wrt~to the reference SfM reconstruction provided by the Cambridge Landmarks datasets~\cite{kendall2015posenet} and Dubrovnik dataset~\cite{li2010location}. We also measured the sizes of the reconstructions using file-system utilities. We considered that an image was successfully registered when it had at least 12 inliers. The baselines for this experiment include state-of-the-art-feature-based localization methods~\cite{sattler2016efficient,svarm2016city,torii201524,zhang2006image}, deep-learning-based ones~\cite{kendall2016modelling,kendall2017geometric,kendall2015posenet,walch2017image}, and state-of-the-art scene-compression algorithms~\cite{camposeco2018hybrid,cao2014minimal,li2010location}. Table~\ref{tab:posenet_datasets} shows the results of this experiment on the Cambridge Landmarks datasets.

{\noindent \bf Localization accuracy marginally improves with a large number of iterations due to our good initialization procedure.}~Fig.\ref{fig:acc_vs_iterations} shows the progression of the rotation and position errors for no more than a million iterations on Old Hospital and King's College. The rotation errors have improvements that are less than one  degree while the position errors have improvements in the order of centimeters.

{\noindent \bf Our method reduces the reconstruction sizes without sacrificing localization accuracy.} Most experiments achieved $100\%$ registration rate of all query images with the exception of the Hybrid compression~\cite{camposeco2019hybrid} that achieved $98\%$ registration rate and Active search~\cite{sattler2016efficient} that achieved $99\%$ registration rate on Old Hospital dataset. Table~\ref{tab:posenet_datasets} shows the sizes of the compressed reconstructions (MB Used columns) and the position and rotation errors (Errors columns). The last nine rows of Table~\ref{tab:posenet_datasets} come from Camposeco~\etal~\cite{camposeco2019hybrid} and use different feature quantization methods, pose estimator algorithms, and a different structure-from-motion library. Thus, the sizes in MB of the compressed reconstructions and pose errors of the last nine rows are not strictly comparable. The first four rows of the Table show the results of our method with different visual distinctiveness scores: 1) initial feasible solution using the avg. descriptor distance score; 2) our method refining the result of 1); 3) fraction of number of cameras per point; 4) and fraction of number of cameras per point normalized with the max. number of cameras seen a point in the reconstruction. The fifth and sixth rows show the k-cover (KC)~\cite{li2010location}, and probabilistic k-cover (KCP)~\cite{cao2014minimal}, respectively. The seventh row shows the localization results using a non-compressed scene. All methods aim to compress at 5\% using Theia library. For KC and KCP, we computed the parameter $K$ that produced a compression close to 5\% but not less. We can observe that our method achieves a similar localization accuracy compared to the accuracy of the baselines in all the datasets while reducing its storage footprint. Table~\ref{tab:dubrovnik_aachen} presents the compression sizes and localization performance on the Dubrovnik dataset; all the compression methods kept around 10\% of the 3D points of the full SfM point cloud. For this experiment, our method used the avg. descriptor distance visual distinctiveness score. The last column in Table~\ref{tab:dubrovnik_aachen} shows the size of the original point cloud and localization performance without compression. We see that our method successfully reduces the size of a large-scale point cloud without sacrificing localization performance. Moreover, our method is efficient and simpler to implement than KC, KCP, and~\cite{camposeco2019hybrid}. However, the good performance of~\cite{camposeco2019hybrid} is due to its modified RANSAC algorithm, its weighted set cover problem, and the two sets of points as described in Sec.~\ref{sec:rel_work}. Unlike~\cite{camposeco2019hybrid}, our method only requires computing a visual distinctiveness score for each point, minimize the proposed cost function via Algorithm~\ref{alg:compression}, and use classical RANSAC and pose estimators.

\vspace{-9mm}
\section{Conclusions}
\label{sec:conclusions}
\vspace{-2mm}
We presented a simple and efficient method that compresses SfM point clouds and keeps the image-based localization accurate. Unlike the K-cover-based methods~\cite{camposeco2018hybrid,cao2014minimal,li2010location}, our method operates by solving a convex quadratic program (QP) that aims to keep 3D points that present a good visual distinctiveness and that maintain a sufficient distance of each other. Our QP formulation resembles the one of one-class support vector machines (SVMs)~\cite{scholkopf2000support}. Given the resemblance with the SVMs, our method keeps the points labeled as support vectors and derived an efficient and easy-to-implement QP solver based on the sequential minimal optimization algorithm~\cite{platt1998sequential}. Our experiments on small- and medium-scale datasets showed that our method operates 20$\times$ faster than ~\cite{cao2014minimal,li2010location} thanks to our computed initial point, and reduces the size of a point cloud by a factor of 35x without sacrificing localization performance.

\noindent {\bf Acknowledgements:} This work was partially funded by NSF grant IIS-1657179.



\appendix
\section{Complete Mathematical Derivation}

We consider the simple constrained QP solver, the following constrained convex optimization problem
\begin{mini!}|l|[2]                   
    {\boldsymbol{\alpha}}                               
    {J(\boldsymbol{\alpha}) \label{eq:objective_min}}   
    {\label{eq:General_QP}}             
    {}                                
\addConstraint{\sum_{i=1}^{m}{\alpha_i}}{=1 \label{eq:constraint_sum}}   
    \addConstraint{0}{\leq \alpha_i \leq \dfrac{1}{\nu m} \quad i=1,...,m  \label{eq:constraint_box}}  
\end{mini!}
where $\boldsymbol{\alpha}$ is a probability distribution over $m$ 3D points, and the parameter $\nu \in \Big[\dfrac{1}{m},1\Big]$ is a factor that controls the compression and it can be fixed a \textit{priori}. This parameter characterizes the solution as follows: \emph{a)} if $\nu$ is lower the sparser the vector $\alpha$ is; and \emph{b)} the higher $\nu$ is the denser $\alpha$ is. The objective function $J(\boldsymbol{\alpha})$ is composed by two terms: the spatial distance term $C$ and the visual distinctiveness $D$. The objective function we aim to minimize is the following
\begin{equation}
J(\boldsymbol{\alpha}) = C - \tau D,
\label{eq:J_cover_dist}
\end{equation}
where $\tau$ controls the spatial distance and visual distinctiveness contributions to the main objective function. The larger the value of $\tau$, the more emphasis on the visual distinctiveness term. As stated in the main submission, our cost function is the following:
\begin{equation}
J(\boldsymbol{\alpha})= \mathbf{\boldsymbol{\alpha}}^T K \mathbf{\boldsymbol{\alpha}} - \tau \mathbf{d}^T \mathbf{\boldsymbol{\alpha}},
\label{eq:J_alpha}
\end{equation}
where $\mathbf{d} \in \mathbb{R}^l$ a vector containing the aggregated confidence score, $\tau$ is the spatial distance-and-distinctiveness trade-off, and $K$ is a kernel matrix.

We used the Gaussian Radial Base function (RBF) to encode the distance among the points:
\begin{equation}
K(\mathbf{x}_i, \mathbf{x}_j)= \exp{\left(-\frac{\|\mathbf{x}_i - \mathbf{x}_j \|^2}{2 \sigma^2}\right)},
\label{eq:rbf}
\end{equation}
where $\sigma \in {\rm I\!R}$ is a kernel parameter and $\|\mathbf{x}_i - \mathbf{x}_j \|^2$ is the Euclidean distance. For convenience, we write $K(\mathbf{x}_i, \mathbf{x}_j)=k_{ij}$ referring to the RBF kernel on the i-th and j-th point.

The sequential minimal optimization (SMO) algorithm aims to find the smallest problem that we can sequentially solve to eventually find the overall solution. To do so, SMO focuses on finding the smallest problem on two variables only. To this end, we rewrite  Eq.\eqref{eq:J_alpha} for $J(\boldsymbol{\alpha})$ as follows:
\begin{equation}
J(\boldsymbol{\alpha})= \sum_{i=1}^{m}\sum_{j=1}^m \alpha_i \alpha_j k_{ij} - \tau \sum_{i=1}^{m} d_i \alpha_i.
\label{eq:J_sum}
\end{equation}

SMO solves the QP problem for a pair of probabilities (\ie, $\alpha_1$ and $\alpha_2$) at each iteration. Without loss of generality, we refer to the first variable with subscript 1 and the second one with subscript 2. Thus, rewriting the Eq.~\eqref{eq:J_sum} emphasizing the pair of variables $(\alpha_1, \alpha_2)$ yields the following expression:
\begin{equation}
\begin{split}
J(\alpha_1,\alpha_2)  = & \sum_{i,j=1}^{2}\alpha_i \alpha_j k_{ij} +  \sum_{i=1}^{2}\sum_{j=3}^m \alpha_i \alpha_j k_{ij} \\ &+ \sum_{i=3}^{m}\sum_{j=3}^m \alpha_i \alpha_j k_{ij} - \tau d_1\alpha_1 -\tau d_2\alpha_2 \\ &- \tau \sum_{i=3}^{m} d_i \alpha_i
\label{eq:J_alpha12}
\end{split}.
\end{equation}

From Eq.~\eqref{eq:J_alpha12} we first note that the spatial distance term is the sum of the all the elements in the following matrix:
\begin{equation}
\begin{bmatrix}
    \alpha_1^2       & \alpha_1\alpha_2k_{12} &  \dots & \alpha_1\alpha_m k_{1m} \\
    \alpha_2\alpha_1k_{21}       & \alpha_2^2 & \dots & \alpha_2\alpha_mk_{2m} \\
    \vdots & \vdots  & \ddots & \vdots \\
    \alpha_m\alpha_1k_{m1}       &\alpha_m\alpha_2k_{m2} & \dots & \alpha_m^2
\end{bmatrix},
\label{eq:matrix_coverage}
\end{equation}
where $\alpha_i$ is the $i^{th}$ alpha entry for the $i^{th}$ point. Using the fact that $k_{ij} = k_{ji}$, we can re-write Eq.\eqref{eq:J_alpha12} as follows:
\begin{equation}
\begin{split}
J(\alpha_1,\alpha_2)  = &  \alpha_1^2+2\alpha_1\alpha_2k_{12}+\alpha_2^2  +2\alpha_1\sum_{j=3}^m \alpha_{j}k_{1j} \\ & + 2\alpha_2\sum_{j=3}^m \alpha_{j}k_{2j} + \sum_{i=3}^{m}\sum_{j=3}^m \alpha_i \alpha_j k_{ij} \\ &    - \tau d_1\alpha_1 -\tau d_2\alpha_2 - \tau \sum_{i=3}^{m} d_i \alpha_i
\label{eq:J_alpha12_coverage}
\end{split}
\end{equation}
where $\sum_{i=3}\sum_{j=3}\alpha_{i}\alpha_{j}k_{ij}$ and $\sum_{i=3}^m C_i \alpha _i$ are  strictly constant with respect to $\alpha_1$ and $\alpha_2$ when optimizing for the pair. Consequently, they can be discarded since they are independent of $\alpha_1$ and $\alpha_2$.

Thus, the total cost function to minimize emphasizing the two $\alpha_1$ and $\alpha_2$ is:
\begin{equation}
\begin{split}
J(\alpha_1,\alpha_2) = & \alpha_1^2 + 2\alpha_1\alpha_2k_{12} + \alpha_2^2 + 2\alpha_1\theta_1 + 2\alpha_2\theta_2 \\ &  - \tau d_1\alpha_1 -\tau d_2\alpha_2,
\label{eq:J_cost_fuction}
\end{split}
\end{equation}
where $\theta_1=\sum_{j=3}^m \alpha_{j}k_{1j}$ and $\theta_2=\sum_{j=3}^m \alpha_{j}k_{2j}$.

\begin{figure}[t]
    \centering
    \includegraphics[width=0.35\textwidth]{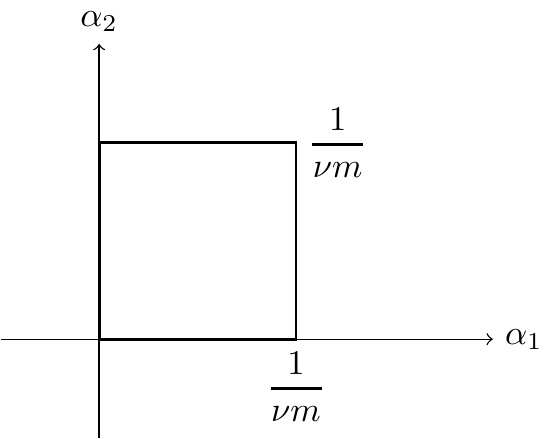}
    \caption{Inequality constraint for the pair $(\alpha_1, \alpha_2)$.}
    \label{fig:optimization_case}
\end{figure}
Using the constraint from Eq.\eqref{eq:constraint_box}, then the pair $(\alpha_1, \alpha_2)$ must lie within a boxed area, as shown in Fig.~\ref{fig:optimization_case}. Also, the two variables must fulfill the equality constraint Eq. \eqref{eq:constraint_sum}. Then, from the probability simplex, \ie, 
\begin{equation}
    \alpha_1 + \alpha_2 + \sum_{i=3}^m = 1,
\label{eq:laphas_sum_constraint}
\end{equation}
we get the following relationships:
\begin{equation}
	\alpha_1+\alpha_2=\Delta ,\quad\quad\quad \textrm{where } \Delta=1-\sum_{i=3}^m \alpha_i.
\label{eq:Delta}
\end{equation}

Given these relationships, we can focus only on the following subproblem:
\begin{mini!}|l|[2]                   
    {\alpha_1, \alpha_2}                               
    {J(\alpha_1, \alpha_2) \label{eq:objective_min}}   
    {\label{eq:QP_alphas}}             
    {}                                
    \addConstraint{\alpha_1+\alpha_2}{=\Delta \label{eq:constraint_sum_alphas}}    
    \addConstraint{0}{\leq \alpha_i \leq \dfrac{1}{\nu m} \quad i=1,2. \label{eq:constraint_box_alphas}}  
\end{mini!}

Using the above linear constrain Eq.\eqref{eq:constraint_sum_alphas} to get $\alpha_2$ as a function of $\alpha_1$ ($\alpha_2=\Delta-\alpha_1$), we can rewrite $J$ as a function of $\alpha_1$ by substituting $\alpha_2=\Delta-\alpha_1$. Thus Eq.\eqref{eq:J_cost_fuction} can be written as:
\begin{equation}
\begin{split}
J(\alpha_1) =& \alpha_1^2 + 2\alpha_1 (\Delta - \alpha_1) k_{12} + (\Delta- \alpha_1)^2 \\&
    +  2 \alpha_1 \theta_1 +  2 (\Delta - \alpha_1)  \theta_2  \\& - \tau d_1 \alpha_1  - \tau d_2 (\Delta - \alpha_1).
\label{eq:J_alpha1}
\end{split}
\end{equation}

Solving the problem at the optimal point, \ie, 
\begin{equation}
\begin{split}
\frac{\partial J}{\partial \alpha_1} &=  2\alpha_1 + 2(\Delta - 2\alpha_1) k_{12} -2(\Delta- \alpha_1)~~ + \\ & ~~~~~~ 2\theta_1 - 2\theta_2  - \tau d_1 + \tau d_2 \\
 &= 0,
\end{split}
\end{equation}
and then solving for $\alpha_1$, we get
\begin{equation}
{\alpha_1}_{\text{optimal}} = \frac{1}{2} \left(\frac{T}{2 (1 - k_{12})}  + \Delta \right)
\end{equation}
where $T=\tau (d_1 - d_2)-2\theta_1 +2\theta_2$.

Since we are solving a box constrained problem we have that
\begin{equation}
    \alpha_1 = \text{max}\left(0, \text{min}\left(\text{min}\left(\frac{1}{\nu m}, \Delta\right), {\alpha_1}_{\text{optimal}}\right)\right).
\end{equation}
Consequently, 
\begin{equation}
\alpha_2 = \Delta - \alpha_1.
\end{equation}

\section{Dataset Statistics}

\begin{table}[t]
    \centering
    \caption{Statistics of the datasets used in the experiments.}
    \footnotesize{
    \begin{tabular}{l >{\centering}p{1.1cm} >{\centering}p{1.1cm} p{1.1cm}}
        \toprule
        Dataset & \# of DB Images & \# of 3D  Points & \# Query Images \\
        \midrule
         King’s College~\cite{kendall2015posenet} & 1,220 & 503K & 343 \\
         Old Hospital~\cite{kendall2015posenet} & 895 & 308K & 182 \\
         Shop Facade~\cite{kendall2015posenet} & 231 & 82K & 103\\
         St. Mary's Church~\cite{kendall2015posenet} & 1,487 & 667K & 530 \\
         Dubrovnik~\cite{li2010location} & 6,044 & 1.2M & 684\\
        \bottomrule
    \end{tabular}
    }
    \label{tab:datasets_supp}
\end{table}

In this section, we aim to provide the statistics of the datasets we used in our experiments. Table~\ref{tab:datasets_supp} shows the statistics of these datasets.

\section{Visual Distinctiveness Scores}

In this section we describe the visual distinctiveness functions that we considered in the main submission.

\subsection{Average Descriptor Distance}

The visual distinctiveness function aims to assign a high score for those 3D points that have a small average descriptor distance from their feature matches, and a low score otherwise. This scoring function thus uses the SfM tracks from the reconstruction to identify the 2D correspondences that are associated to the 3D point. Given the $l$-th 3D point and their corresponding 2D match pairs $\left(i, j\right) \in \mathcal{P}_l$, the score uses the following function:
\begin{equation}
    d_l = \exp\left( -{\frac{1}{|\mathcal{P}_l| \beta} \sum_{\left(i, j\right) \in \mathcal{P}} \| \mathbf{y}_{i} - \mathbf{y}_j \|} \right),
    \label{eq:matchability}
\end{equation}
where $\mathbf{y}_i$ and $\mathbf{y_j}$ are the SIFT descriptors from the $\left(i, j\right)$ correspondence pair, $\mathcal{P}_l$ is the set of correspondences of the $l$-th point, and $\beta$ is a normalization parameter. The exponential function returns a value close to $1$ when the average descriptor distance is near zero and returns $0$ when the average distance is large. Thus, $d_l$ tends to $1$ for points that have descriptors that are close to each other in feature space, and tends to $0$ otherwise.

\subsection{Frequency-based Functions}
The next method of assigning a score to every point is based upon the frequency of that point being seen across images. For example, a higher score will be assigned to a point that is seen from 50 cameras than a point seen from 10 cameras. We use two different measures for this frequency. The first is simply the fraction of cameras the point is seen in. The second is the fraction of cameras the point is seen in \wrt~ the maximum number of cameras that see one point in the reconstruction.

\subsubsection{Fraction of Number of Images a Point is Seen In}
This score is the ratio between the number of cameras that depict the $i$-th point. A point that is seen from 0 cameras (technically not possible, it would not exist in the reconstruction otherwise), would be assigned a score of $0$. A point that is seen from every camera would be assigned a score of $1$.  This is represented by
\begin{equation}
    d_i = \frac{n_i}{N},
    \label{eq:freqnumpoints}
\end{equation}
where the number of cameras seeing the $i$-th point is $n_i$, and $N$ is the number of total cameras in the reconstruction. Thus, this score function will favor points that are seen in numerous images. Intuitively, this means that the 3D point is visually distinctive since many images depict it.

\subsubsection{Fraction of Number of Images a Point is Seen In Normalized By The Maximum Number of Cameras a Point is Seen In}
This score is similar to the one shown in Eq.\eqref{eq:freqnumpoints}. However, the normalization value $N = \max\left\{ n_1, \hdots, n_m \right\}$. In other words this score assigns the distinctiveness value of the current point \wrt~the maximum number of cameras that see a point in the reconstruction. For example, if the best point is seen within 50 views, it assigns a score of $1$ to it. All other scores are proportional to the number cameras viewing it out of 50. Mathematically, this is expressed as follows:
\begin{equation}
    \text{score} = \frac{n_i}{\text{max}\left\{n_1,...,n_m\right\}},
    \label{eq:freqbestpoint}
\end{equation}
where $n_i$ is the number of cameras that see the $i$-th point and $m$ is the number of points.

\subsection{Combination.}
The combination based score is computed as a convex combination between distance and frequency based scores. The weight is a tunable parameter, but for the purpose of our results, we used 0.5. This score is simply,
\begin{equation}
    \text{score} = w \cdot d + (1-w) \cdot f
    \label{eq:combination}
\end{equation}
where $w \in \left[0, 1\right]$ is the weight, $d$ is the distance score computed in \eqref{eq:matchability}, and $f$ is the frequency score computed in \eqref{eq:freqbestpoint}.

\subsection{Performance of Visual Distinctiveness Functions}

\begin{table*}[t]
    \centering
    \caption{Localization performance and compression results on the Cambridge Landmarks datasets~\cite{kendall2015posenet} using different visual distinctiveness scores. The Table presents the size in MBs of the compressed reconstructions (MB Used columns) and position and orientation errors (Errors columns). Despite using different visual distinctiveness scores (first four rows), our methods reduce the SfM point cloud the most compared to KC~\cite{li2010location} and KCP~\cite{cao2014minimal}. Our methods achieve a comparable localization accuracy \wrt~to the baselines despite using a smaller SfM point cloud.}
    \label{tab:posenet_datasets_distinctiveness}
    \scriptsize{
    \setlength{\tabcolsep}{1.8pt} 
    \begin{tabular}{l >{\centering}p{0.8cm} >{\centering}p{1.4cm} >{\centering}p{0.8cm} >{\centering}p{1.4cm} >{\centering}p{0.8cm} >{\centering}p{1.4cm} >{\centering}p{0.8cm} p{1.4cm}}
    \toprule
    \multirow{2}{*}{\bf Method} & \multicolumn{2}{c}{\bf King’s College}& \multicolumn{2}{c}{\bf Old Hospital} & \multicolumn{2}{c}{\bf Shop Facade} & \multicolumn{2}{c}{\bf St. Mary's Ch.} \\ 
     & MB Used & Errors $\left[m, \circ\right]$ & MB Used & Errors $\left[m, \circ\right]$ & MB Used & Errors $\left[m, \circ\right]$ & MB Used &  Errors $\left[m, \circ\right]$ \\
     \midrule
     {\bf Avg. Distance} & 2.2 & 1.53, 1.09 & 1.1 & 0.90, 2.17 & 0.41 & 0.72, 1.40 & 3.3 & 0.56, 0.89\\
     {\bf Num. Images} & 2.2 & 0.94, 0.57 & 1.1 & 1.33, 0.99 & 0.41 & 0.31, 0.56 & 3.3 & 0.43, 0.64\\
     {\bf Best Point} & 2.2 & 0.92, 0.59 & 1.1 & 1.24, 0.96 & 0.41 & 0.31, 0.59 & 3.3 & 0.46, 0.65 \\
     {\bf Combination} & 2.2 & 0.89, 1.05 & 1.1 & 1.67, 1.18 & 0.41 & 0.55, 1.05 & 3.3 & 0.44, 0.65\\
     \midrule
     KC~\cite{li2010location} & 3.1 & 1.48, 1.23 & 6.0 & 1.35, 1.06 & 0.85 & 0.51, 0.87 & 18 & 0.46, 0.69\\
     KCP~\cite{cao2014minimal} & 5.9 & 0.99, 0.86 & 8.2 & 1.19, 1.00 & 1.3 & 0.44, 0.80 & 24 & 0.40, 0.61\\
     No comp. & 98 & 0.57, 0.50 & 34 & 0.96, 0.79 & 11 & 0.23, 0.43 & 131 & 0.28, 0.45\\
     \bottomrule
    \end{tabular}
    }
\end{table*}

\begin{table*}[t]
    \centering
    \caption{Localization performance results on the NotreDame dataset~\cite{snavely2006photo} using different visual distinctiveness scores. The Table presents position and orientation errors (Errors columns). The distinctiveness score that returned the lowest position and rotation errors is the avg. distance visual distinctiveness score. The position errors in this dataset are not in metric scale.}
    \label{tab:notredame_datasets_distinctiveness}
    \begin{tabular}{l c c}
    \toprule
     & Rotation Error [deg]& Position Error\\
    \midrule
    Avg. Distance & 0.58 & 5.86\\
    Num. Images & 0.71 & 8.23 \\
    Best Point & 0.66 & 7.46 \\
    Combination & 0.64 & 6.29\\
    \bottomrule
    \end{tabular}
\end{table*}

We analyze the localization and compression performance of our method when using the visual distinctiveness functions discussed in the previous section. The results of these experiments on the Cambridge Landmarks dataset are shown in Table~\ref{tab:posenet_datasets_distinctiveness} and results on NotreDame dataset~\cite{snavely2006photo} are shown in Table~\ref{tab:notredame_datasets_distinctiveness}. We can observe in Table~\ref{tab:posenet_datasets_distinctiveness} that frequency-based and combination scores achieve good localization errors while keeping a good compression rate. On the other hand, we observe that the avg. distance score produce good localization errors on the NotreDame dataset while frequency-based and combination scores produced the worse localization errors compared to the avg. distance score.

\section{Visualization of Compressed Scenes}

\begin{figure*}[t!]
    \centering
    \includegraphics[width=\textwidth]{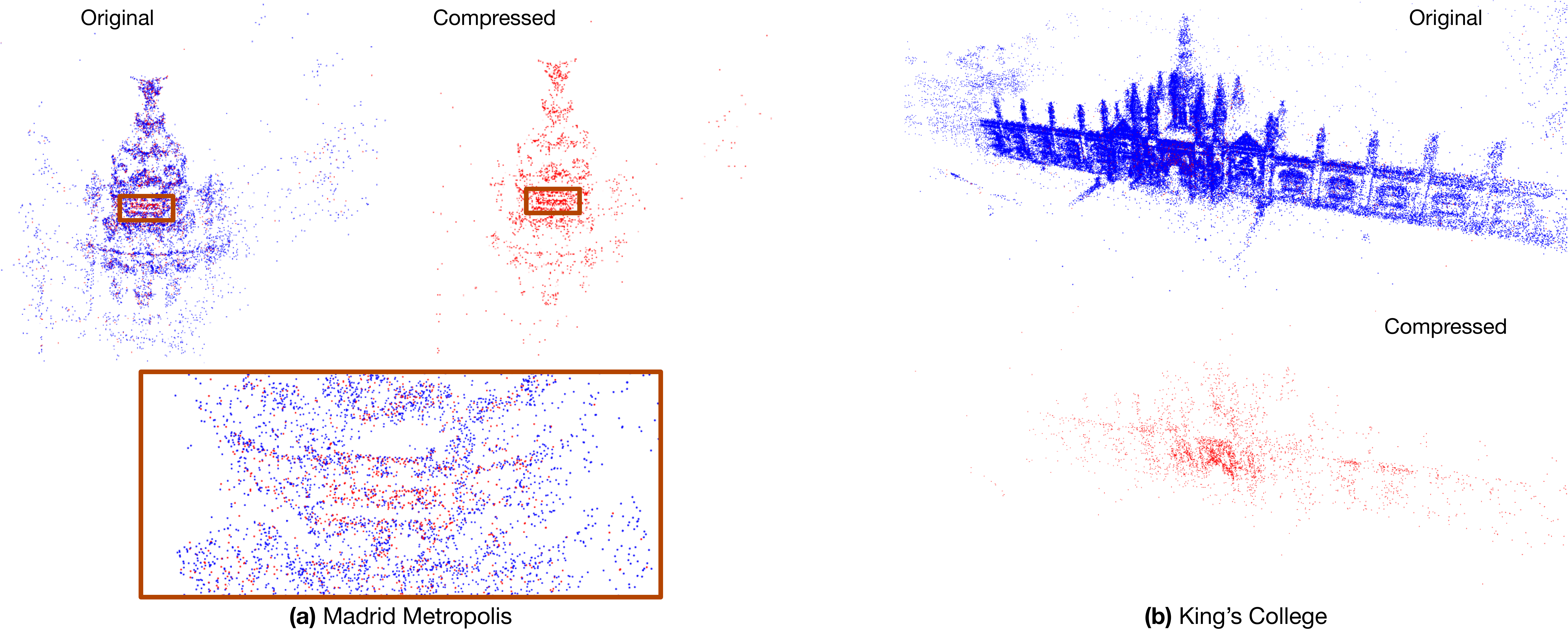}
    \caption{Visualization of original SfM reconstructions ({\color{blue} blue}) and compressed scene representations ({\color{red} red}). {\bf (a)} Visualization of a section of Madrid Metropolis dataset. {\bf (b)} Visualization of King's College dataset.}
    \label{fig:visualizations}
\end{figure*}

Fig.~\ref{fig:visualizations} in the supplemental material shows two original SfM reconstructions (shown in blue) and their compressed scene representations (shown in red) using our proposed algorithm. We can observe in Fig.~\ref{fig:visualizations}(a) that our algorithm selected points from the dome and facade, including points detected on the text on a sign at the top of the building. Also, we can see in Fig.~\ref{fig:visualizations}(b) that our algorithm mainly kept points on the facade of the King's College building.

{\footnotesize
\bibliographystyle{ieee}
\bibliography{references}
}

\end{document}